\documentclass[runningheads]{llncs}

\usepackage{cite}
\usepackage{amsmath,amssymb,amsfonts}
\usepackage{algorithmic}
\usepackage{graphicx}
\usepackage{textcomp}
\usepackage{subfig}
\usepackage{xcolor}
\usepackage{upgreek}
\usepackage{soul}
\usepackage{float}

\usepackage{todonotes}

\def\BibTeX{{\rm B\kern-.05em{\sc i\kern-.025em b}\kern-.08em
    T\kern-.1667em\lower.7ex\hbox{E}\kern-.125emX}}
\begin{document}

\title{Visualising Evolution History in Multi- and Many-Objective Optimisation}
\titlerunning{Visualising Evolution History in Multi- and Many-Objective Optimisation}
\author{Mathew J. Walter \and David J. Walker \and Matthew J. Craven}
\institute{School of Engineering, Computing and Mathematics,\\University of Plymouth,\\ United Kingdom, PL4 8AA.\\ \email{\{mathew.walter,david.walker,matthew.craven\}@plymouth.ac.uk}}

\maketitle

\begin{abstract}
Evolutionary algorithms are widely used to solve optimisation problems. However, challenges of transparency arise in both visualising the processes of an optimiser operating through a problem and understanding the problem features produced from many-objective problems, where comprehending four or more spatial dimensions is difficult. This work considers the visualisation of a population as an optimisation process executes. We have adapted an existing visualisation technique to multi- and many-objective problem data, enabling a user to visualise the EA processes and identify specific problem characteristics and thus providing a greater understanding of the problem landscape. This is particularly valuable if the problem landscape is unknown, contains unknown features or is a many-objective problem. We have shown how using this framework is effective on a suite of multi- and many-objective benchmark test problems, optimising them with NSGA-II and NSGA-III.

\keywords{Visualisation \and Evolutionary computation \and Multi-objective optimisation}
\end{abstract}


\section{Introduction}
Optimisation problems abound in science and industry, and in recent decades a plethora of approaches have arisen to solve them. A prominent example are \textit{evolutionary algorithms} (EAs). An EA takes an initial population and uses nature-inspired operators to perturb the solutions towards optimal solution (or solutions). As well as solving an optimisation problem, it is important that the processes with which they are generated are understandable by non-expert problem owners, and often this is not the case: therein lies a challenge of transparency. Visualisation is a natural approach to addressing this issue, exposing the solutions, and the mechanisms used to generate them, to the end user.

This paper expands upon \cite{de2019analysis}, which compared the extent to which  dimension reduction techniques preserved population movements and the exploration-exploitation trade-off. It also proposes two compact visualisations, one of which we extend to visualise the search history of an EA optimising a multi- and many-objective problem. We use this to identify specific characteristics of problems as well as identifying the population dynamic through the evolution process. This paper offers the following novel contributions:

\begin{enumerate}
    \item The method is applied to a suite of multi- and many-objective test problems containing a wider set of problem features that can be visualised than the single-objective problems used in \cite{de2019analysis}.
    \item Specific problem features that can be be identified through the  proposed visualisation are identified and examples are shown.
    \item The method is applied in both the search space and objective space. Specific problem characteristics and population movements can be more prominent in a low-dimensional embedding of a particular space (i.e. discontinuities in the search space may only be able to be identified in the objective space embedding visualisation for a particular problem).
\end{enumerate}

The remainder of this paper is structured as follows: In Section 2, relevant formal definitions are introduced. We review existing work on visualising many/multi-objective optimisation from the literature, and we acknowledge the paper  
this work extends. Section 3 contains details of the dimension reduction techniques, the test problem suite, problem features and a summary of the methodology implemented for visualisation. The experimental setup, containing details of the parameters used in this experiment are highlighted in Section 4. Section 5 hosts the results and analysis and provides a discussion of the many-objective problems. We 
conclude and discuss future work in Section 6.

\section{Background}
\label{Background}
A multi-objective optimisation problem comprises $M$ competing objectives, such that a solution $\boldsymbol{x}$ is quantified by an objective vector $\boldsymbol{y}$ with $M$ elements:
\begin{equation}
    \begin{aligned}
     \boldsymbol{y} = (f_1(\boldsymbol{x}), \ldots, f_M(\boldsymbol{x}))\\
      \textrm{such that} \: \boldsymbol{x} \in \Omega, \: \boldsymbol{y} \in \Lambda
    \end{aligned}
\end{equation}
where $\Omega$ is the \textit{search space} and $\Lambda$ is the \textit{objective space.} Many-objective optimisation problem comprises four or more competing objectives (and thus $M \geq 4$ for such problems).
The task of a multi-objective evolutionary algorithm (MOEA) and many-objective evolutionary algorithm (MaOEA) is to optimise a problem comprising a set of $M$ conflicting objectives to which there can be no solution that simultaneously optimises all $M$ objectives. Solutions are compared using the dominance relation, whereby solution $\boldsymbol{y}_i$ dominates solution $\boldsymbol{y}_j$ if it is no worse than $\boldsymbol{y}_j$ on any objective and better on at least one. More formally, assuming a minimisation problem without loss of generality:
\begin{equation}
    \boldsymbol{y}_i \prec \boldsymbol{y}_j \Longleftrightarrow \forall m(y_{im} \leq y_{jm}) \wedge \exists m(y_{im} \textless y_{jm}).
\end{equation}
If neither $\boldsymbol{y}_i$ dominates $\boldsymbol{y}_j$, nor $\boldsymbol{y}_j$ dominates $\boldsymbol{y}_i$, then the solutions are {\it mutually non-dominating}. A solution with no dominating solutions is {\it non-dominated}. The goal of a MaOEA/MOEA is to identify the Pareto set, the set of feasible solutions that cannot be dominated. The objective space image is called the Pareto front.

\subsection{Previous visualising search history literature}

Rather than visualising the Pareto set and/or Pareto front, visualising the search population can generate more useful information to the decision-maker (DM). The DM will often be interested not only in the non-dominated solutions but also the mechanism from which they are generated. The existing literature concerning visualising population movements in evolutionary multi-objective optimisation (EMO) is minimal. One study that does consider population movement presents a visual method for benchmarking the performance of EAs. The method is used to illustrate a range of good and bad performance characteristics \cite{walker2018toward, walker2020identifying}. Other examples of visual methods for examining algorithm parameters are \cite{craven2014ea, Kobayashi2016APO}. Existing work from \cite{brockhoff2017quantitative} visualises search history in EMO.%

Other papers are concerned with visualising the non-dominated solutions \cite{walker2012visualizing,6777535}. However, these do not address a significant issue: the DM's comprehension of the algorithm's population movements. 
Further work that visualises the whole population is that of \cite{he2015visualization}, which maps individuals from a high-dimensional objective space into a 2-D polar coordinate plot while preserving the Pareto dominance relationship. Like the studies referenced above, that work is concerned with configuring MaOEAs rather than characterising the problem landscape.

There are many choices of content to be visualised. For example, one could visualise the search space, the objective space or search process. This work considers visualising all three with multi/many-objective problems. As shown in Section \ref{Results}, often more information can be captured when visualising both spaces simultaneously than by visualising a single space. Further, this work chooses to visualise the whole population rather than a subset (such as the Pareto front) to capture the maximum amount of population movement information from the visualisation. This paper pays particular attention to visualising the population as the population evolves through the algorithm.

\section{Visualising Search History}
\label{VSH}

One of the difficulties that arise from visualising many-objective optimisation data is being able to comprehend four or more spatial dimensions visually. 
In order to make a visualisation tool applicable to the class of multi/many-objective EAs, a dimension reduction technique needs to be applied to the population. The mapping $\Uppi$ is a dimension reduction mapping if $\Uppi : \mathbb{R}^m \rightarrow \mathbb{R}^n$, where $n \textless m$. In this work, Multi-Dimensional Scaling (MDS)\cite{torgerson1952multidimensional} is the chosen technique for dimension reduction; this is due to its effectiveness at maintaining population structure \cite{de2019analysis}. MDS is used to translate pairwise distances of the population individuals into a lower-dimension Cartesian space.

As discussed in Section \ref{Background}, this work chooses to visualise the search space and the objective space. Through visualising the population evolving through the space, one is able to identify specific problem features. There are many choices of problem features that can be considered. In this work, we consider four:
\begin{itemize}
\item Local optimum - A local optimum is a solution that is optimal within a neighbouring set of candidate solutions.

\item Modality - For the problems considered, the objective functions can be unimodal or multimodal. An objective function is unimodal if it has a single optimum, or multimodal if it has multiple local optima.

\item Bias - A problem is biased if there's significant density variation of solutions in the objective space, given an even spread of solutions in parameter space.

\item Disconnected Pareto optimum set/front - In this case, a problem has a Pareto set/front in disconnected regions.
\end{itemize}

In order to detect these problem features, the experiment requires a diverse choice of test problems to allow various properties of the individual problems to be identified in the visualisations; for this work the DTLZ test suite \cite{deb2002scalable} is used. The problems contain many features which can be used to support the exposure of population movements. The application of the DTLZ test suite allows one to identify specific problem features from the population MDS - however, the DTLZ test suite is far from comprehensive. It is noted in \cite{huband2006review} that the DTLZ test suite has several limitations such as: none of its problems feature fitness landscapes with flat regions, none of its problems are deceptive, none of its problems are (practically) nonseparable and the number of position parameters is always fixed relative to the number of objectives.

\subsection{Visualising Search History Methodology}
\label{VSH METHOD}
The visualisation method used herein is based on that defined by \cite{de2019analysis}, visualising the search history of an optimiser once the optimisation process has completed. The optimiser results in a sequence of populations in which $P_i$ is the population from the $i$-th generation, ranked according to its members' fitness values, where $i \in \{1,\ldots, n_{gen}\}$, where $n_{gen}$ is the total number of generations. This sequence of populations is concatenated into a single multiset, the dimensionality of which is reduced using multidimensional scaling (MDS) from $M$ to 2. In all cases herein, $M>2$. The resulting embedding is then used for visualisation, with the two embedded coordinates forming the $x$ and $y$ coordinates, and the generation number providing the value for the $z$-axis.

Within the visualisation, colour is used to illustrate the trade-off between search and exploitation, showing which mode of optimisation the algorithm is currently operating in. The work of
\cite{vcrepinvsek2013exploration,de2019analysis} is employed to determine to what level the set of all solutions at a particular generation is being explored or exploited. Exploration is inversely proportional to exploitation. This metric is applied to the visualisation in \cite{de2019analysis}.

The exploration and exploitation metric is calculated for each generation. At each generation, the Euclidean distance between each pairwise individual in the population is calculated, and the minimum distance is saved from which the median minimum pairwise distance is calculated. At each generation, the minimum distance for each pairwise individual is compared against the overall median. Thus the individuals with lower distance are considered to be exploiting.

\section{Experimental Setup}

The experiment comprises of running EAs on five continuous problems from the DTLZ test suite\cite{deb2002scalable}, namely, DTLZ1-4 and DTLZ7. These five problems have real-valued decision variables lying in the region $[0,1]$. The suggested number of decision variables is $D=k+M-1$, where $k = 5$ for DTLZ1, $k=20$ for DTLZ7, and $k = 10$ for the remaining problems. The problems are scalable in the number of objectives; in this experiment, three and five objective problems are utilised. The 3-objective problems are optimised with NSGA-II \cite{deb2002fast} and the 5-objective problems with NSGA-III \cite{deb2013evolutionary}. The crossover probability is 0.8, and the mutation probability is set to 0.1. The distribution index, controlling the size of the perturbation, in both cases is fixed (15 for SBX, 7 for polynomial mutation). The algorithm's runtime is 100,000 function evaluations for $M=3$ and 200,000 for $M=5$. The visualisations are then produced as in Section \ref{VSH METHOD}.

Having generated data by optimising one of the problems with either NSGA-II or NSGA-III, the search and objective spaces are visualised.
The MDS plots of both whole populations in the objective and search spaces are generated.
The points in the visualisations are coloured according to the exploration-exploitation metric, except, however, the final generation in each visualisation which is plotted as a white cross.
An analysis is then performed of the plots, with the intention of both identifying characteristics of the test problems and identifying the dynamics of the population as it evolves through the problem.

\section{Results} 
\label{Results}
\subsection{Multi-objective Problems} 

\begin{figure}[t]
    \centering
    \subfloat{{\includegraphics[width=.57\textwidth, clip=true, trim=3cm 1.8cm 1.5cm 3cm]{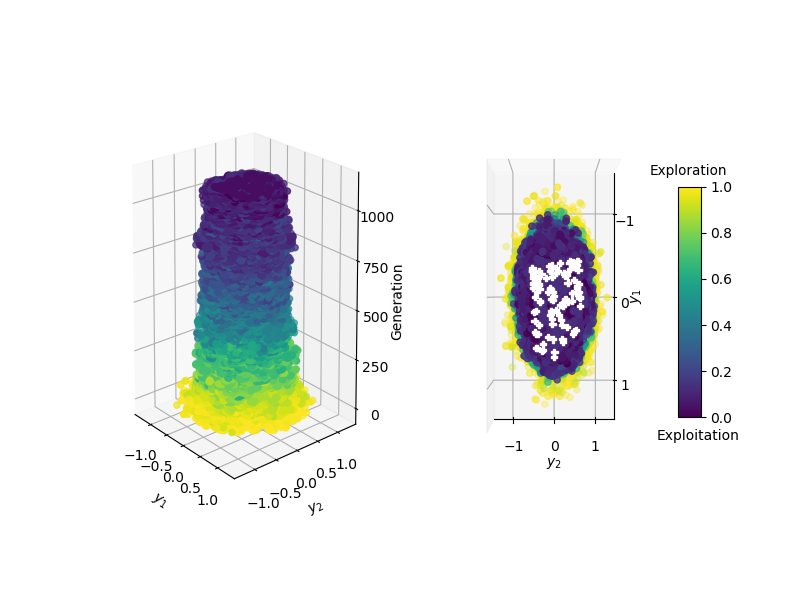}} }
    \qquad 
    \subfloat{{\includegraphics[width=.57\textwidth, clip=true, trim=3cm 1.8cm 1.5cm 3cm]{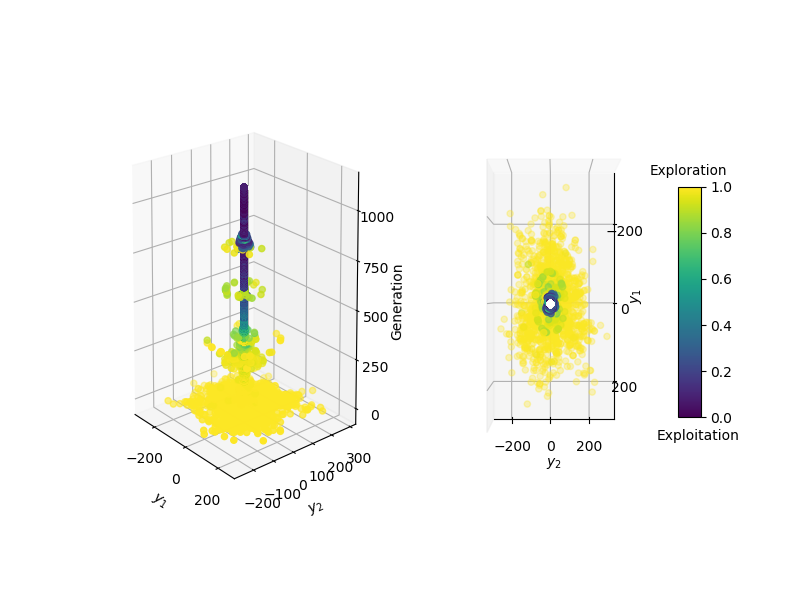}} }
    \qquad
    \vspace{-0.1cm}\caption{DTLZ1, with the top illustrations showing the MDS reduced search space. The bottom illustrations show the MDS reduced objective space. The solutions are coloured according to their exploration-exploitation metric.}
    \label{fig:DTLZ1}
\end{figure}

Figure \ref{fig:DTLZ1} illustrates results for optimising DTLZ1 in three objectives. The top panel visualises the search space, and the lower panel shows the corresponding objective space results. DTLZ1 is multimodal, and the effect of this on the search process can be seen in the result.
As the population approaches optima, the population movements should decrease; by definition, this decreases the exploration measure. The MDS mapping would translate a decrease in population movements to an MDS population confined together and converged around $(0,0)$ in the $(y_1,y_2)$ plane. From this perspective, the population distances can be observed to converge to approximately $(0,0)$ when finally reaching the global optima (this occurs in the final 250 generations in the objective space). However, before the objective space population converges to the global optimum, a subset of the population (identified as the `rings' in Figure \ref{fig:DTLZ1}) are exploring after diverging from a local optimum, whilst the remaining population are converging to the global optimum. This can be observed at approximately generations 250, 500 and 750. Furthermore, the population is coloured according to the exploration-exploitation metric. At the identified local optima, the population posses lighter colours such as yellow and green, suggesting the points are exploring in an attempt to escape from the local optima.

\newpage

\vspace*{-1.0cm}
The final generation presents some structure from the final generation in the objective space (a triangular plane). This corresponds to earlier work using MDS to visualise many-objective populations, wherein it was shown to preserve the structure and, to an extent, geometry of a mutually non-dominating set \cite{walker2012visualizing}. As in both spaces, the exploration/exploitation seems to indicate that as the generations increase, exploration increases and hence exploitation decreases; this matches intuition.
\begin{figure}[t]
    \centering
    \subfloat{{\includegraphics[width=.57\textwidth, clip=true, trim=3cm 1.8cm 1.5cm 3cm]{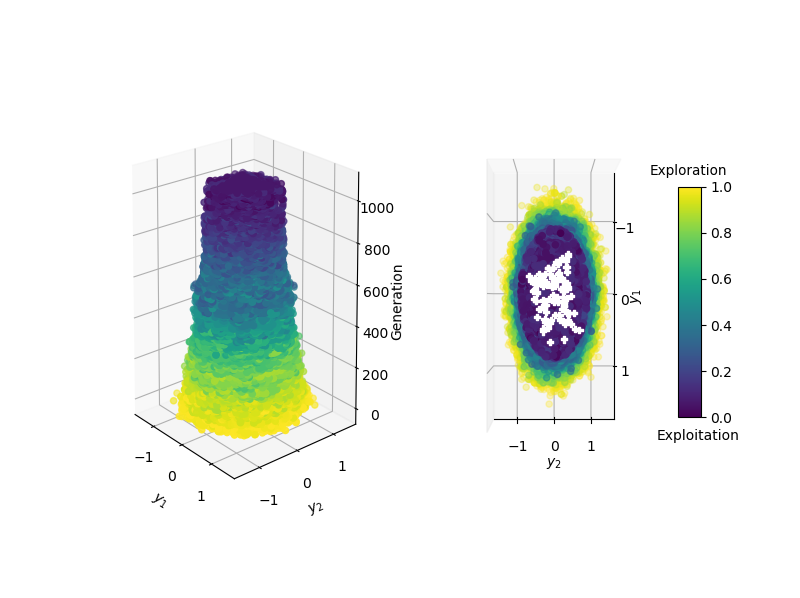} }}%
    \qquad
    \subfloat{{\includegraphics[width=.57\textwidth, clip=true, trim=3cm 1.8cm 1.5cm 3cm]{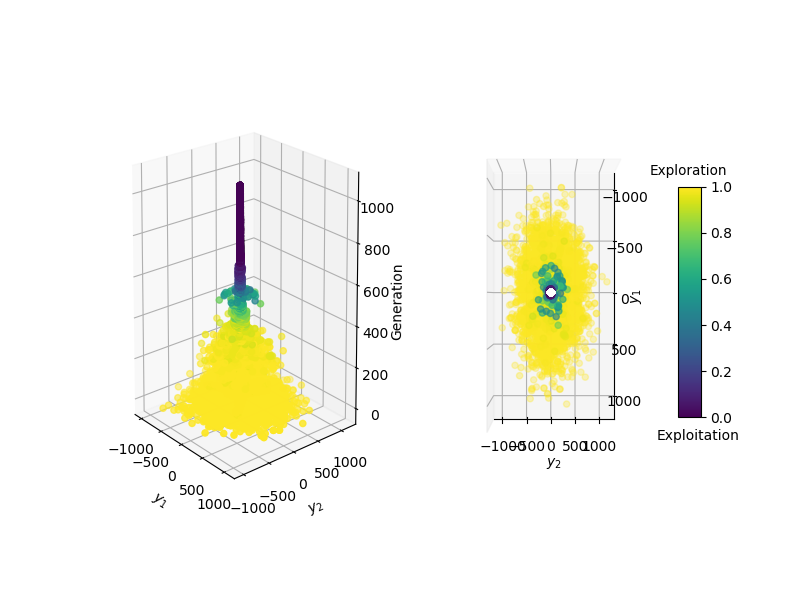}} }%
    \vspace{-0.1cm}\caption{DTLZ3, the top illustrations showing the MDS reduced search space. The bottom illustrations show the MDS reduced objective space.}
    \label{fig:DTLZ3}%
\end{figure}

DTLZ3 is also a multimodal problem, and similar results can be observed in Figure \ref{fig:DTLZ3} to that of DTLZ1. In the MDS reduced search space the population appears to converge, with the population distances reducing as the optimiser evolves. At around generation 400, it can be seen to diverge and converge again as the population encounters a local minimum. On the MDS reduced search space the final generation preserves some of the structure from the final generation in the objective space (the positive orthant of the unit sphere).

\begin{figure}[t] 
    \centering
    \subfloat{{\includegraphics[width=.57\textwidth, clip=true, trim=3cm 1.8cm 1.5cm 3cm]{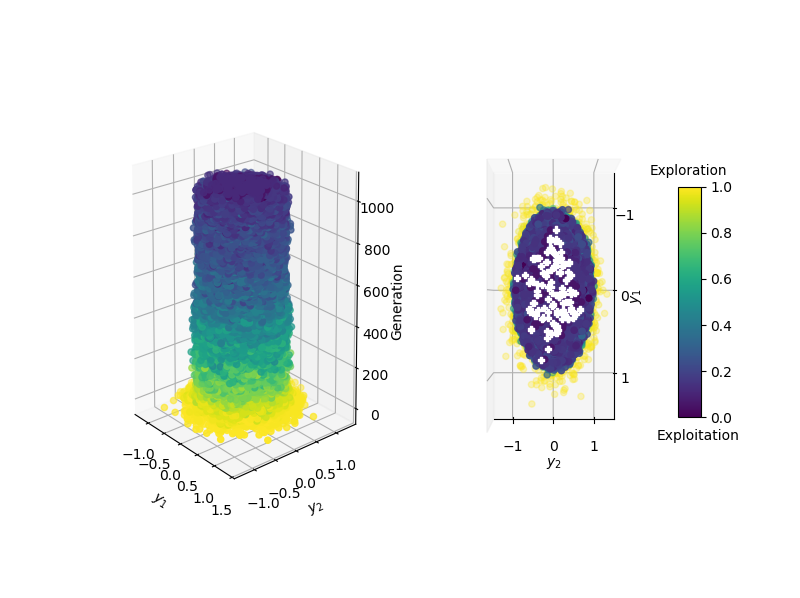} }}%
    \qquad 
    \subfloat{{\includegraphics[width=.57\textwidth, clip=true, trim=3cm 1.8cm 1.5cm 3cm]{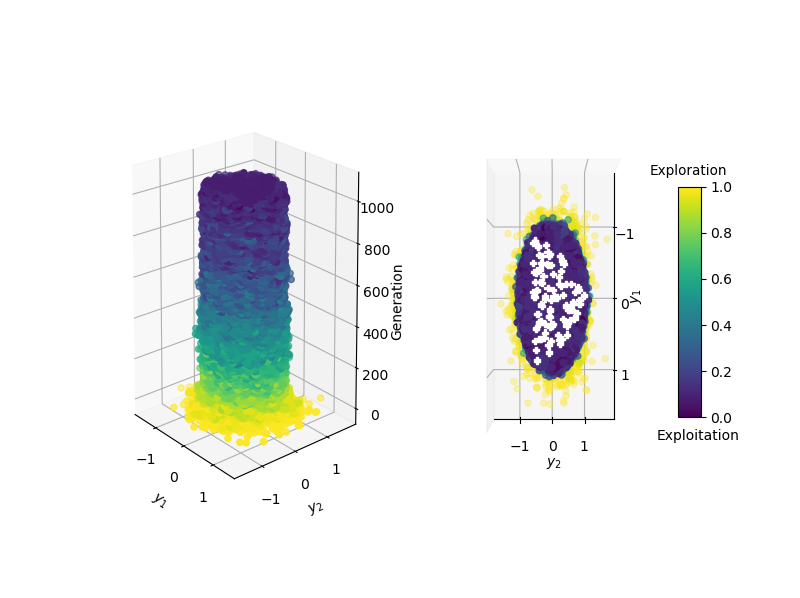}} }%
    \vspace{-0.2cm}\caption{DTLZ2, the top illustrations showing the MDS reduced search space. The bottom illustrations show the MDS reduced objective space.}
    \label{fig:DTLZ2}%
\end{figure}

\begin{figure}[t!]
    \label{DTLZ4}
    \centering
    \subfloat{{\includegraphics[width=.57\textwidth, clip=true, trim=3cm 1.8cm 1.5cm 3cm]{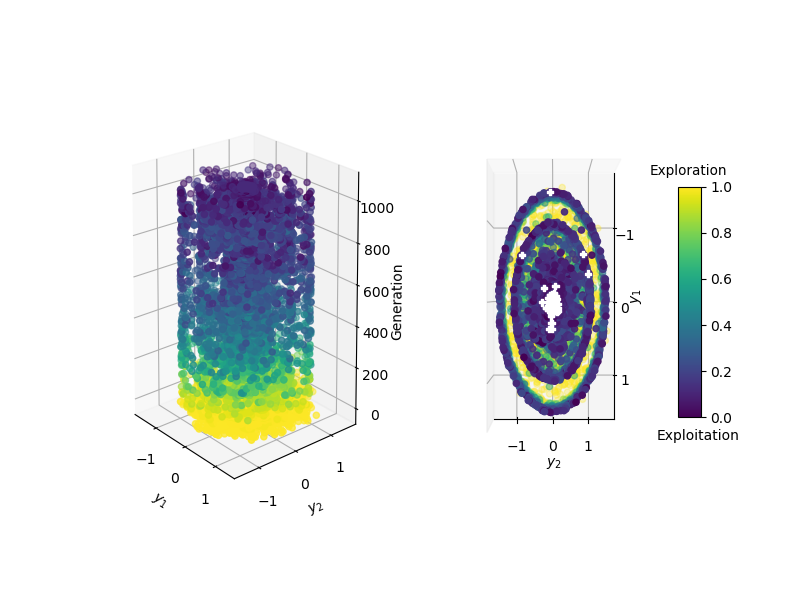} }}%
    \vspace{-0.2cm}\caption{DTLZ4, showing the MDS reduced search space.}
    \label{fig:DTLZ4}%
\end{figure}

DTLZ2 and DTLZ4 are illustrated in Figures \ref{fig:DTLZ2} and \ref{fig:DTLZ4} respectively. Both problems are unimodal, and there is usually much less to see as the problems are less challenging. 
For DTLZ2, both spaces appear to be cylindrical shapes with very little character; this is because the problems contain a simple search space and there is little to prevent an optimiser converging to the global optimum very quickly. This is reflected in the spaces. The final generation in the MDS space has preserved much of the final generation structure in the objective and search space. The objective space of DTLZ2 is similar to that of DTLZ3 and DTLZ4, and hence the final generations are all similar in structure. The transition from population exploration to exploitation is more gradual than the exploration-exploitation transition in the multimodal problems; this appears intuitive.

In the visualisations illustrated so far, it seems most of the information is contained within the MDS reduced objective space. In the case of DTLZ4 there appear to be more interesting characteristics of the plot in the MDS reduced search space, this can be seen in Figure \ref{fig:DTLZ4}. In the decision space MDS the population appears to form circular `shockwaves'. This is because the search space contains a dense area of solutions next to the $f_{M}/f_{1}$ plane.  DTLZ4 is a biased problem, which increases the difficulty of a problem by making it harder to converge to and fully cover the Pareto front.

\begin{figure}[t]
\vspace*{-0.5cm}
\centering
    \subfloat[
        Search space
    ]{{\includegraphics[scale=0.3]{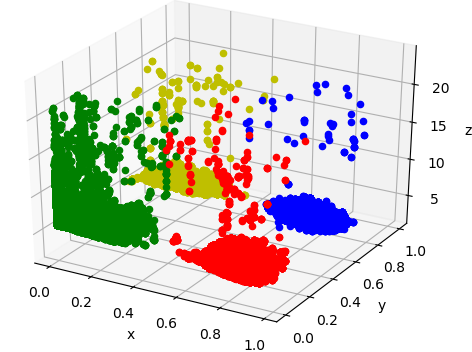}} }%
    \subfloat[
        Search space search history
    ]{{{\includegraphics[scale=0.3, clip=true]{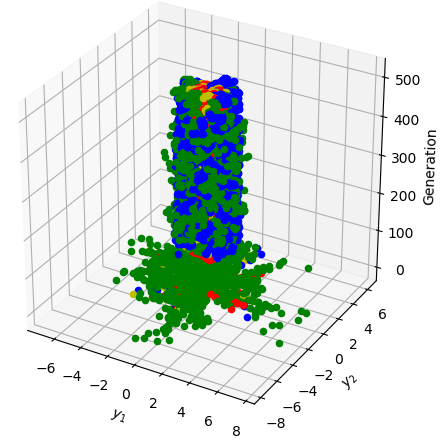}}} }%
    \hspace{0.4cm}\subfloat[
            Final generation
        ]
        {{\includegraphics[scale=0.19]{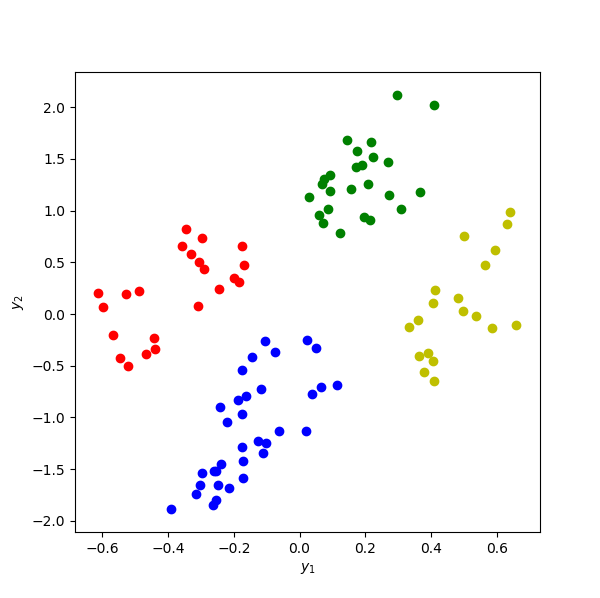} }}
    \label{2DFinalGen}
    \vspace{-0.1cm}\caption{Clustering coloured MDS reduced search space. For (a), axes $x,y,z$ correspond to the three objectives. In (b) and (c), axes $y_i$ correspond to the reduced MDS data axes.} %
        \label{fig:DTLZColour}%
\end{figure}

\vspace{-0.06cm} 
DTLZ7 is a mixed modality problem. Objectives  $f_{1:M-1}$
are unimodal and objective $f_{M}$ is multimodal. This problem has disconnected Pareto-optimal regions in the search space. There appears to be more similarity with the unimodal objective problems.
The disconnected regions of the Pareto front are shown in Figure \ref{fig:DTLZColour}, and can be seen as they evolve through the search space. Note, the regions are clearly visible in the MDS reduced search space embedding of the final generation. This structure is not visible in the objective space visualisation, and this problem is an example of a case in which considering both spaces can yield useful information (in this case the objective space visualisation yields no new information, and is omitted due to lack of space).

\newpage
\begin{figure}[t!]
    \centering
    \subfloat{{\includegraphics[width=.57\textwidth, clip=true, trim=3cm 1.8cm 1.5cm 3cm]{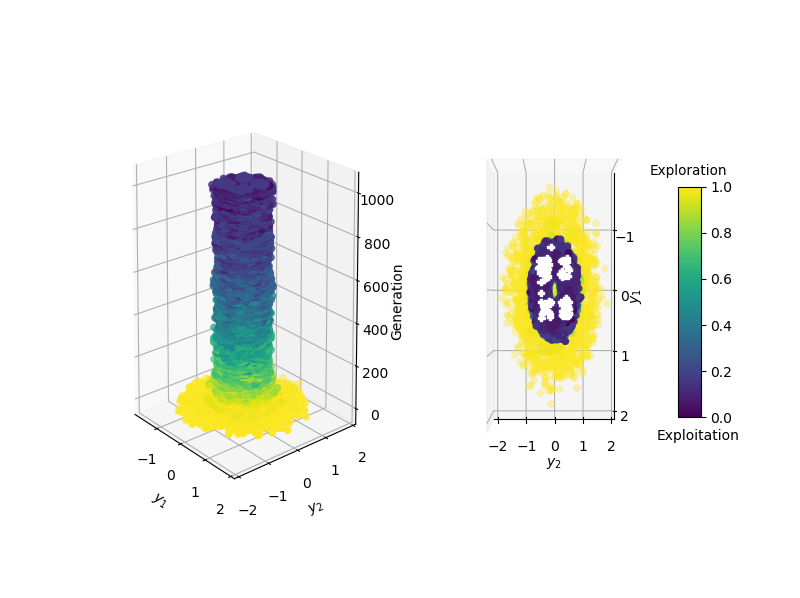} }}%
    \vspace{-0.1cm}\caption{DTLZ7, showing the MDS reduced search space.}
    \label{fig:DTLZ7}%
\end{figure}

\vspace{-0.2cm}
\begin{figure}[t]
\vspace{-0.5cm}
\centering
  \subfloat{\raisebox{0.7cm}{\includegraphics[width=.35\textwidth, clip=true, trim=.25cm .1cm 1cm 1cm]{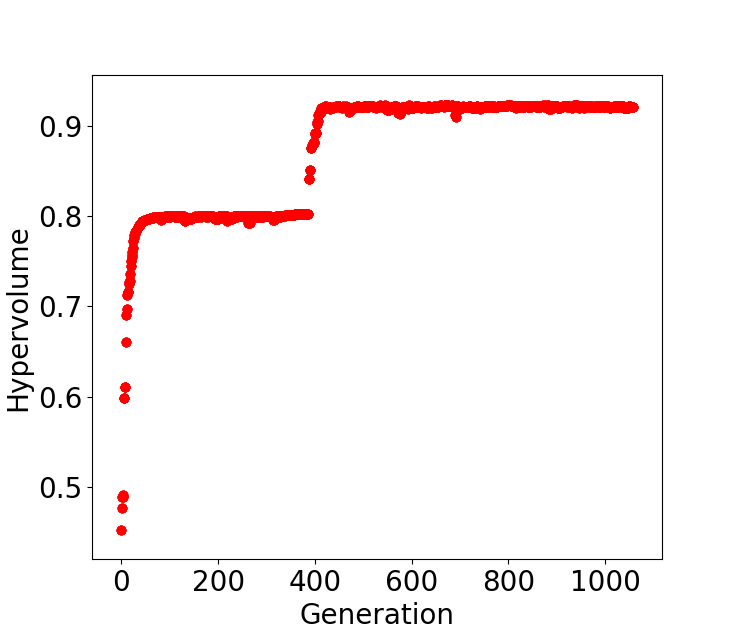}}} \hfill
    \subfloat{{\includegraphics[width=.57\textwidth, clip=true, trim=3cm 1.8cm 1.5cm 3cm]{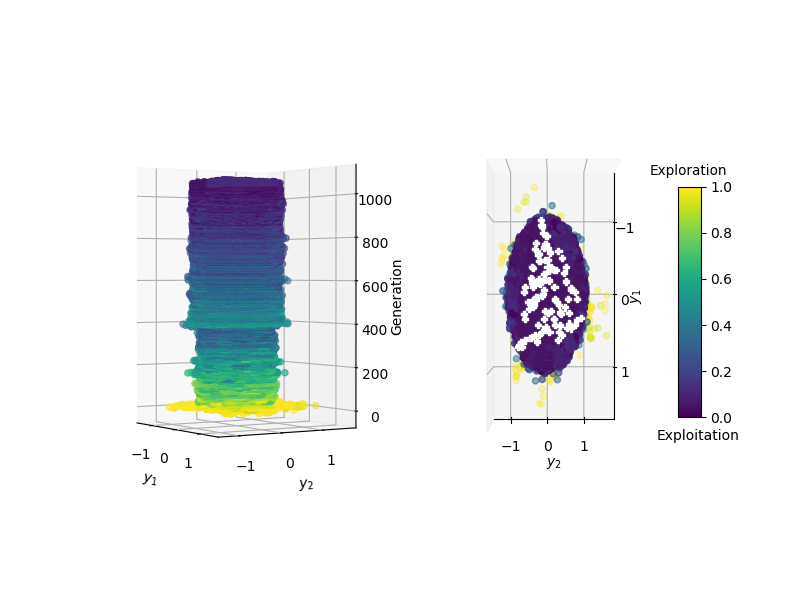}} }
   \vspace{-0.4cm}
   \caption{The hypervolume of a three objective DTLZ4 problem, followed by the corresponding MDS reduced objective space plot.}
  \label{fig:HyperVolume}
\end{figure}

\vspace*{-0.8cm}
Figure \ref{fig:HyperVolume} illustrates the use of the hypervolume indicator \cite{fleischer2003measure} in parallel with the MDS visualisation in order to understand population movements. The hypervolume is a measure of performance widely used to assess the progress of a MOEA in terms of both convergence and diversity -- in order to achieve the maximum possible hypervolume score the Pareto front approximation must converge to the Pareto front and cover it completely. In this case, the population appears to converge, in approximately the first 30 generations. Then up to around 400 generations, the population distances are small; subsequently, the population seems to converge to the optima. These changes in optimiser progress correspond to the changes that are visible in the objective space visualisation, indicating that they are highlighting the same artefact in the optimisation history.

\vspace{-0.2cm}
\subsection{Many-objective Problems}
The framework is demonstrated on five objective problems, for which a larger number of function evaluations, and hence generations, are required. In this experiment, 200,000 function evaluations are run.

\begin{figure}[t]
    \centering
    \subfloat{{\includegraphics[width=.57\textwidth, clip=true, trim=3cm 1.8cm 1.5cm 3cm]{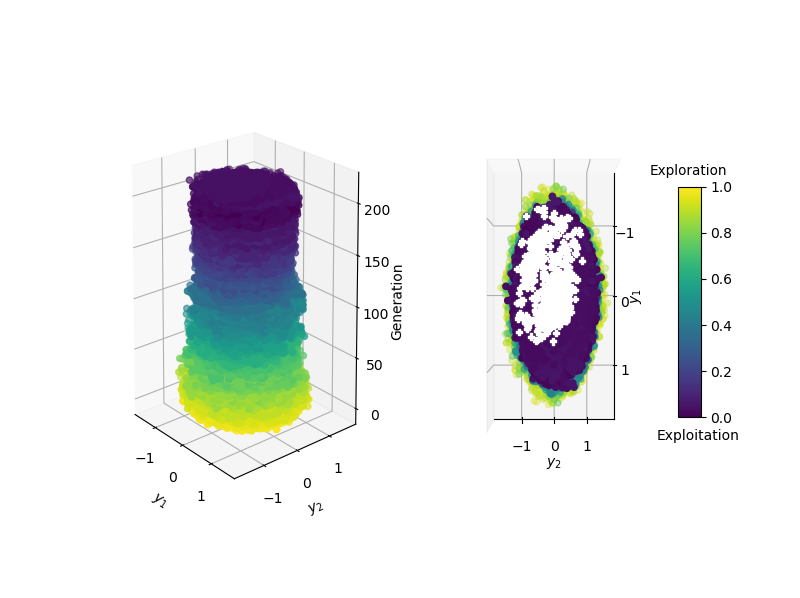} }}%
    \qquad
    \subfloat{{\includegraphics[width=.57\textwidth, clip=true, trim=3cm 1.8cm 1.5cm 3cm]{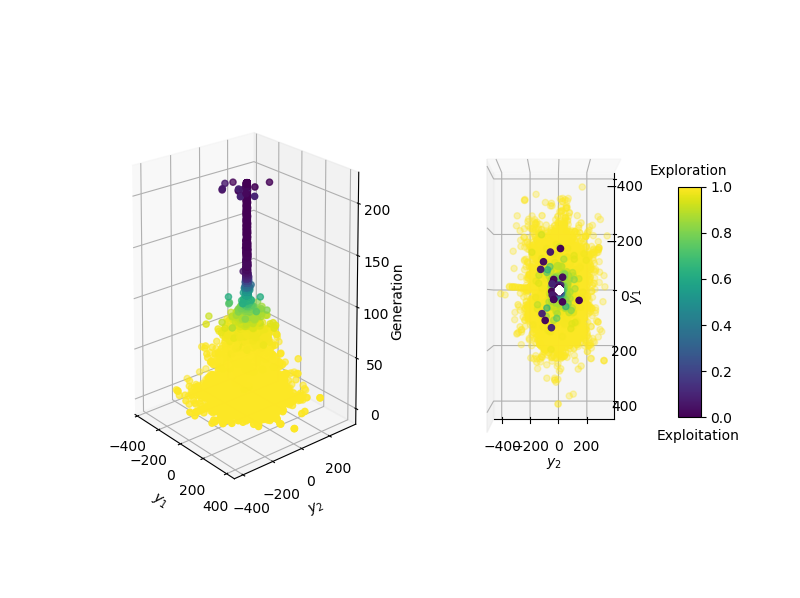}} }%
    \vspace{-0.1cm}\caption{DTLZ1 with five objectives, top illustrations showing the MDS reduced search space. The bottom illustrations show the MDS reduced objective space.}
    \label{fig:DTLZ1-5D}%
\end{figure}

\begin{figure}[h!]
    \centering
    \qquad
    \subfloat{{\includegraphics[width=.57\textwidth, clip=true, trim=3cm 1.8cm 1.5cm 3cm]{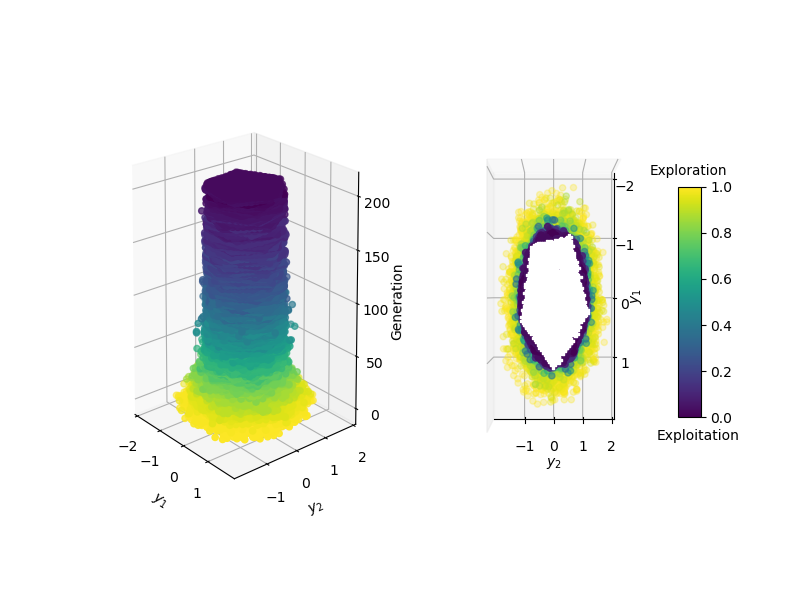}} }%
    \vspace{-0.1cm}\caption{DTLZ2 with five objectives, showing the MDS reduced objective space.}
    \label{fig:DTLZ2-5D}%
\end{figure}

\begin{figure}[t!]
    \centering
    \subfloat{{\includegraphics[width=.57\textwidth, clip=true, trim=3cm 1.8cm 1.5cm 3cm]{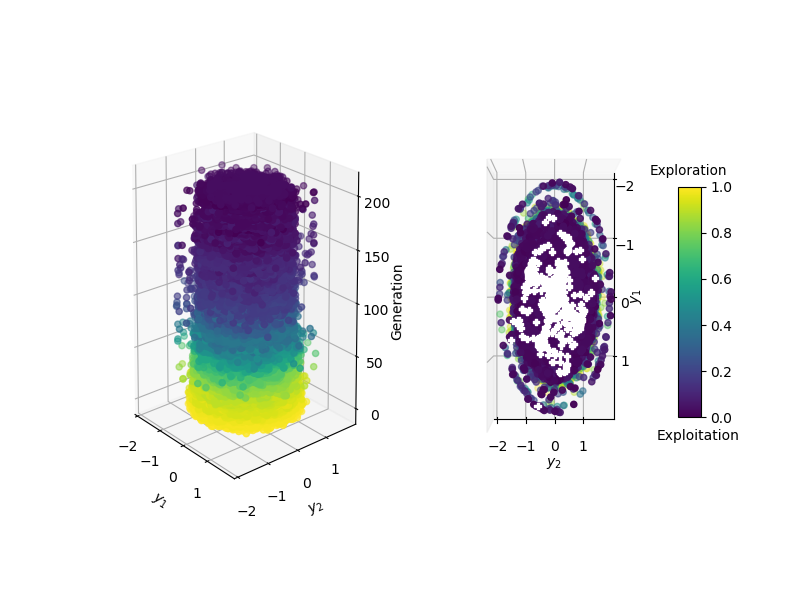} }}%
    \vspace{-0.1cm}\caption{DTLZ4 with five objectives, showing the MDS reduced search space.}
    \label{fig:DTLZ4-5D}%
\end{figure}

\begin{figure}[t!]
    \centering
    \qquad
    \subfloat{{\includegraphics[width=.57\textwidth, clip=true, trim=3cm 1.8cm 1.5cm 3cm]{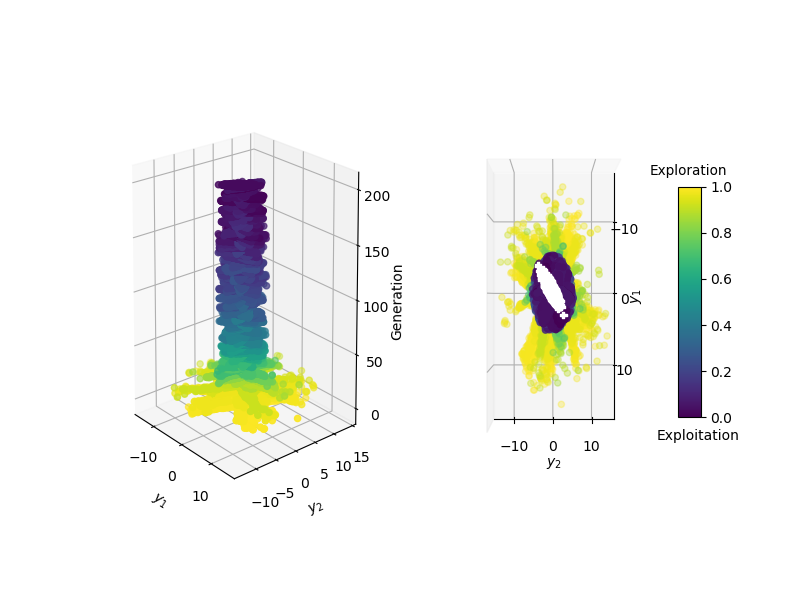}} }%
    \vspace{-0.1cm}\caption{DTLZ7 with five objectives, showing the MDS reduced objective space.
    }%
    \label{fig:DTLZ7-5D}%
\end{figure}

In Figure \ref{fig:DTLZ1-5D}, the DTLZ1 test problem appears to converge around an optimum at approximately generation 100. In the final generations the population diverges again after encountering another optimum; this is because some population distances have increased. We can see the population has not converged as quickly as in the three objective problem; this is intuitive, as the five objective problem is more difficult. 
DTLZ2 (Figure \ref{fig:DTLZ2-5D}) and DTLZ4 (Figure \ref{fig:DTLZ4-5D}) have a final generation which shows a very distinctive pentagon in the MDS reduced objective space, which demonstrates the final generation in the MDS visualisation maintains a similar structure to the final generation in the objective space. It should be noted: the pentagon is formed due to the five objective problem nature. For a problem comprising $M$ objectives, one would expect to find a $M$-sided shape from the final generation MDS reduced objective space. DTLZ4 contains a ring around the MDS reduced search space. The population appears to form circular `shockwaves'. This is because the search space contains a dense area of solutions next to the $f_{M}/f_{1}$ plane. 
In the MDS reduced search space of DTLZ7 (Figure \ref{fig:DTLZ7-5D}), the clusters of points become more difficult to observe than with the same problem in three objectives. The MDS reduced objective space, however, appears to show stacked `lines', and shows how the NSGA-III algorithm operates on the population movements. NSGA-III uses a set of reference directions to maintain diversity among solutions, and the population appears to converge along these reference points. We therefore state, the problem characterisations that can be inferred from the visualisations are highly dependent on the employed algorithm. That is, the visualisations show the search behavior from which the problem characterisations can be seen only indirectly, leading to some visualisation features being an artifact of the algorithm.

\vspace{-0.3cm}
\section{Conclusion}
\label{sec:conclusion}

By extending the visualisation proposed in \cite{de2019analysis}, we have identified specific problem characteristics through population movements that could be of value to the DM.
This provides a better understanding of the problem landscape, and how the algorithm is performing, allowing the DM to make decisions based on how the optimisers are evolving solutions to the problem.
We have shown how this framework can be used to locate some of the characteristics of a problem. For example, the framework has identified clusters caused by discontinuities in the Pareto front, can identify 
local optima, and allows one to see where the population approximately converges to the Pareto front. 
Ultimately, the visualisation illustrates how the population moves through the search space.

The approach can be used to identify important characteristics of a problem; this is particularly useful if the problem landscape is unknown, and contains unknown features. 
It is well known that visualising many-objective solutions is a challenge, and this work has shown how using this framework is effective for both multi- and many-objective problems. 
Work on the proposed method is ongoing, and we are currently examining techniques for further highlighting problem features within the visualisation. This is in addition to considering a wider range of problem features, and other types of problems (e.g., discrete problems and real-world problems).

It is clear how different information is preserved in the the search and objective spaces, and that the both spaces should be used in parallel to maximise the information obtained about the population movements. We are currently exploring interactive visualisations that are based on a linear combination plot of the two, as well as allowing a user to manipulate the combination in an interactive visualisation, allowing them to run through the populations in both the objective/search space and the MDS space simultaneously. Ultimately, the aim of this ongoing work is to help the DM and researchers identify the population movements within their problems and provide a better comprehension of the algorithms and problems. Enabling this kind of transparency within genetic algorithms will make the use of genetic algorithms more accessible to DMs.

\bibliographystyle{plain}
\bibliography{Bibliography}
\end{document}